\pdfoutput=1

\documentclass[11pt]{article}

\usepackage[]{EACL2023}

\usepackage{times}
\usepackage{latexsym}

\usepackage[T1]{fontenc}

\usepackage[utf8]{inputenc}

\usepackage{microtype}

\usepackage{inconsolata}

\usepackage{mathtools}
\usepackage{multirow}
\usepackage{graphicx}
\usepackage{amsfonts}
\usepackage{enumitem}
\usepackage{hhline}
\usepackage{siunitx}

\usepackage{caption}
\usepackage{multicol}

\DeclarePairedDelimiter\ceil{\lceil}{\rceil}

\newcommand*\RETRO{\textsc{Retro}}
\newcommand*\RETROON{\textsc{Retro[on]}}
\newcommand*\RETROOFF{\textsc{Retro[off]}}
\newcommand*{\RET}{\text{\textsc{Ret}}}

\title{On the Generalization Ability of Retrieval-Enhanced Transformers}

\renewcommand{\thefootnote}{\fnsymbol{footnote}}

\def\authorsep{\hspace{0.3em}}

\author{Tobias Norlund$^{1,4}$\thanks{\ \ Corresponding author, \texttt{tobiasno@chalmers.se}} \authorsep Ehsan Doostmohammadi$^{2}$ \authorsep Richard Johansson$^{1,3}$ \authorsep Marco Kuhlmann$^{2}$\medskip\\
\null$^{1}$ Chalmers University of Technology \quad \null$^{2}$ Linköping University\\
\null$^{3}$ University of Gothenburg \quad \null$^{4}$ Recorded Future}

\begin{document}

\renewcommand{\thefootnote}{\arabic{footnote}}

\maketitle

\begin{abstract}
Recent work on the Retrieval-Enhanced Transformer (\RETRO) model has shown that off-loading memory from trainable weights to a retrieval database can significantly improve language modeling and match the performance of non-retrieval models that are an order of magnitude larger in size.
It has been suggested that at least some of this performance gain is due to non-trivial generalization based on both model weights and retrieval.
In this paper, we try to better understand the relative contributions of these two components.
We find that the performance gains from retrieval largely originate from overlapping tokens between the database and the test data, suggesting less non-trivial generalization than previously assumed.
More generally, our results point to the challenges of evaluating the generalization of retrieval-augmented language models such as \RETRO, as even limited token overlap may significantly decrease test-time loss. 
We release our code and model at \small{\texttt{https://github.com/TobiasNorlund/retro}}
\end{abstract}

\section{Introduction}

Large-scale generative language models have shown promising results toward creating a general-purpose foundation for many natural language applications. 
While sheer scale-up has resulted in better language modeling performance, the immense costs are an inhibiting factor towards further improvements \citep{llm_costs}.

Recent work on retrieval-augmented language models, such as the Retrieval-Enhanced Transformer (\RETRO; \citealp{retro}), suggests that \emph{memory} can be effectively off-loaded from the model parameters to an external database.
In \RETRO, the information retrieved from the database is used to augment the context from which the model predicts new tokens, reducing the need to memorize this information in the model parameters.
This opens up for smaller language models with retained performance.
Specifically, \citet{retro} report that, with a large enough retrieval database, \RETRO\ can achieve a performance comparable to GPT-3 \citep{gpt3} and Jurassic-1 \citep{J1WhitePaper} on the Pile \citep{pile}, at only 4\% of the parameters.
Similarly, \RETRO\ achieves significantly lower bits-per-byte performance compared to a baseline of the same size without retrieval.

\citet{retro} conclude that \RETRO\ has the capacity for non-trivial generalization based on both the model parameters and the retrieval database, even though they find that part of the performance gains can be attributed to lexical overlap between retrieval and test data.
In this work, we want to better understand the nature and magnitude of this effect.
Our findings indicate that performance gains\footnote{Results on \RETRO\ were originally reported in bits-per-byte, while we report results in loss.} originate \emph{almost exclusively} from \RETRO's ability to copy tokens verbatim from retrieved data, effectively exploiting any (small or large) overlap between training and test data.
This suggests that the ability of \RETRO\ to fuse retrieved and in-parameter information may be more limited than previously assumed.

\section{Method}

To investigate gains from retrieval, we re-implement the \RETRO\ model described by \citet{retro} (with a few deviations; see below). We present the model here in brevity.

\subsection{The \RETRO\ Model}

\RETRO\ is an autoregressive language model trained with the next-token prediction objective, where the prediction probability is conditioned on additional context retrieved from a database.

\paragraph{Retrieval}
\label{sec:retrieval}

Retrieval occurs at the granularity of contiguous token chunks with a fixed size~$m$.
More specifically, assume that \RETRO\ has already generated a sequence of tokens $x_{1:t}$.
Each token $x_i$ belongs to a chunk $C_{c(i)}$, where $c(i) = \ceil*{i/m}$.
The probability of the next token $x_{t+1}$ depends on the previously generated tokens and the context retrieved from the previously seen chunks:
\begin{displaymath}
    P\left(x_{t+1} \,|\, x_{1:t}, \RET(C_1), \dots, \RET(C_{c(t+1)-1}); \theta \right)
\end{displaymath}

\paragraph{Database}

\RETRO's database takes the form of a key--value storage $R(N) \mapsto [N, F]$, where $N$ is a chunk from one of the indexed documents, $F$ is the immediately following chunk, and the key $R(N) \in \mathbb{R}^d$ is the embedding of~$N$ according to some embedding model $R$.
This database is used to retrieve the $k$ nearest neighbors of a chunk $C$, based on the embedding $R(C)$:
\begin{displaymath}
    \RET(C) = ([N^1, F^1], \dots, [N^k, F^k])
\end{displaymath}

\paragraph{Architecture}

\RETRO\ is based on the original Transformer architecture \citep{transformer}.
Chunk neighbors are encoded by the encoder and attended to by the decoder.
Due to the quadratic complexity in self-attention, each neighbor is encoded separately; all representations are then concatenated and made available to the decoder \citep{izacard-grave-2021-leveraging}.
The original decoder is modified such that for the prediction of token $x_{t+1}$, cross-attention (CA) can only attend to the neighbor representations retrieved based on the previous chunk $C_{c(t+1)-1}$.
This is called \emph{chunked cross-attention} (CCA).
Furthermore, the encoder is modified to include a restricted form of cross-attention to the decoder.
Specifically, the encoder CA attends to the decoder hidden states immediately before the first CCA.
We refer to \citet{retro} for more details.

\paragraph{Implementation Details}

For tokenizing documents, we use the pre-trained T5 tokenizer.
The retrieval was performed using approximate nearest neighbor search with the high-performant \texttt{faiss} library \citep{faiss}.
We implement \textsc{Retro} in PyTorch \citep{pytorch} and use PyTorch Lightning for distributing the training and validation data across GPUs and compute nodes.
Our implementation deviates from that of \citet{retro} only in that we
\begin{itemize}[leftmargin=*, itemsep=0pt]
    \item use learnable relative positional biases as in T5 \cite{raffel2020t5}, with a bucket for each unique relative position; and
    \item instantiate the chunk embedding model $R$ by a pretrained Sentence-BERT (SB) model \citep{reimers-gurevych-2019-sentence} instead of \textsc{Bert}. We deemed SB to be preferable over \textsc{Bert} as it is smaller (i.e.\ cheaper to compute) and produces embeddings of lower dimensionality (i.e.\ saves disk space).
\end{itemize}

\subsection{Dataset}
\label{seq:massiveopentext}

\citet{retro} used a multi-lingual version of \emph{MassiveText} \cite{gopher} for both training and retrieval data.
To replicate the English portion of this data, we sought open-source alternatives.
\emph{MassiveText} comprises text from the categories web text, news, code, books, and Wikipedia.
By pooling matching categories from Pile \citep{pile} and adding the RealNews dataset \citep{realnews}, we obtain a large dataset composed of all five categories, consisting of 36M documents and 52B tokens.
We keep the training/validation splits from the Pile categories.
For RealNews, we use the provided training set and a subsample of 16,400 documents from the validation set.
The full description of our dataset is shown in Table~\ref{tab:massiveopentext}.

\subsection{Model Training}

For our experiments, we train a \RETRO\ model that resembles the 425M model\footnote{The 425M parameters exclude embeddings.} in \citet{retro}, as shown in Table~\ref{tab:models}.
We train and test on our open-source version of \emph{MassiveText} as described in Section~\ref{seq:massiveopentext}.
During training, we retrieve neighbors from the training set, while at validation time, we retrieve from the union of training and validation sets.
We filter out neighbors that originate from the same source document as the query chunk.
Each model is trained on sequences of no more than 1,024 tokens; longer sequences are truncated.
We use a chunk size of 64 and retrieve two neighbors during both training and validation.
We train the model for 140k training steps with a batch size of 16.
This means that only 6\% of the training documents are actually used during training, excluding retrieved neighbors.
We use the Adam optimizer with a fixed learning rate of \num{1e-4}.

\section{Experiments}

\begin{figure}[]
    \centering\vskip-\abovedisplayskip
    \includegraphics[width=0.5\textwidth]{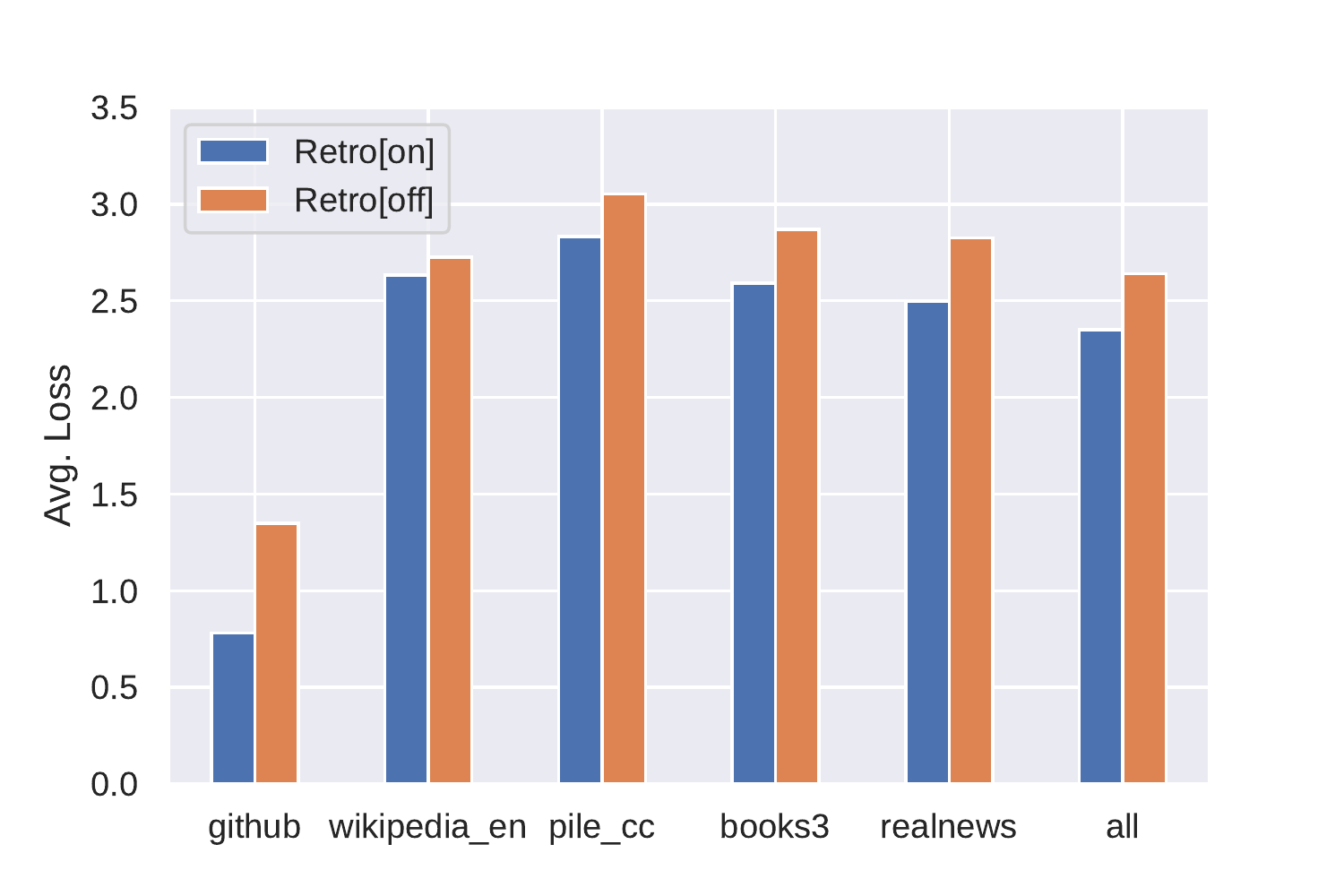}
    \caption{Comparing loss on validation set categories, when using retrieval vs.\ no retrieval.}
    \label{fig:ppl}
\end{figure}

\citet{retro} observed that retrieval increases language modeling performance.
To validate this observation, we compare two configurations of our model: \RETROON, where we enable retrieval, and \RETROOFF, where we remove the CCA layers, thereby reducing \RETRO\ to a standard decoder-only language model.
As we can see in Figure \ref{fig:ppl}, retrieval reduces the loss across all data categories, and with 11\% across the full validation set.
GitHub data has the lowest validation loss among all categories and is also where we see the largest reduction in loss, at 42\%.
Wikipedia sees the smallest reduction in loss, at only 3\%.
A closer comparison to the results from \citet{retro} is available in Appendix \ref{appendix:model_validation}.

\subsection{Loss per Degree of Overlap}
\label{sec:consecutive_overlap}

As \citet{retro} note, retrieval-based models such as \RETRO\ may more easily exploit evaluation dataset leakage.
To quantify how much of the positive effect of retrieval on language modeling performance can be attributed to such leakage, the authors computed bits-per-byte (bpb) for evaluation chunks with different amounts of consecutive token overlap relative to their retrieved neighbors.
This analysis showed that, while the positive effect of retrieval decreased with smaller overlaps, it was still significant at overlap levels of at most 8 contiguous tokens, which the authors considered small enough to conclude that while \RETRO\ actually learns to \emph{generalize} from retrieval data, not merely copy-and-paste it.
Here we investigate the hypothesis that the bpb reductions observed by \citet{retro} \emph{are localized exclusively in the overlapping tokens}.
If this was true, it would challenge the conclusion that \RETRO\ learns non-trivial generalizations based on retrieval data.

To test our hypothesis, we sort the validation set tokens into buckets based on their leftward overlap.
Specifically, we put a token $x_i$ into a bucket $\Phi(n)$, where $n$ is the largest number such that $x_i$ and the $n-1$ tokens preceding it consecutively overlap with some neighboring chunk in $\RET(C_{c(i)-1})$.
For example, the bucket $\Phi(1)$ contains all tokens $x_i$ for which the unigram $x_i$ appears in some neighbor, but not the bigram $x_{i-1} x_i$; the bucket $\Phi(2)$ contains all $x_i$ for which $x_{i-1} x_i$ overlaps but not $x_{i-2} x_{i-1} x_i$, and so on. 
As a special case, the bucket $\Phi(0)$ contains all tokens that do not overlap with any of its neighbors.
This includes all tokens that occur in a first chunk $C_1$, which lacks neighbors.

\begin{figure}[]
    \centering\vskip-\abovedisplayskip
    \includegraphics[width=0.5\textwidth]{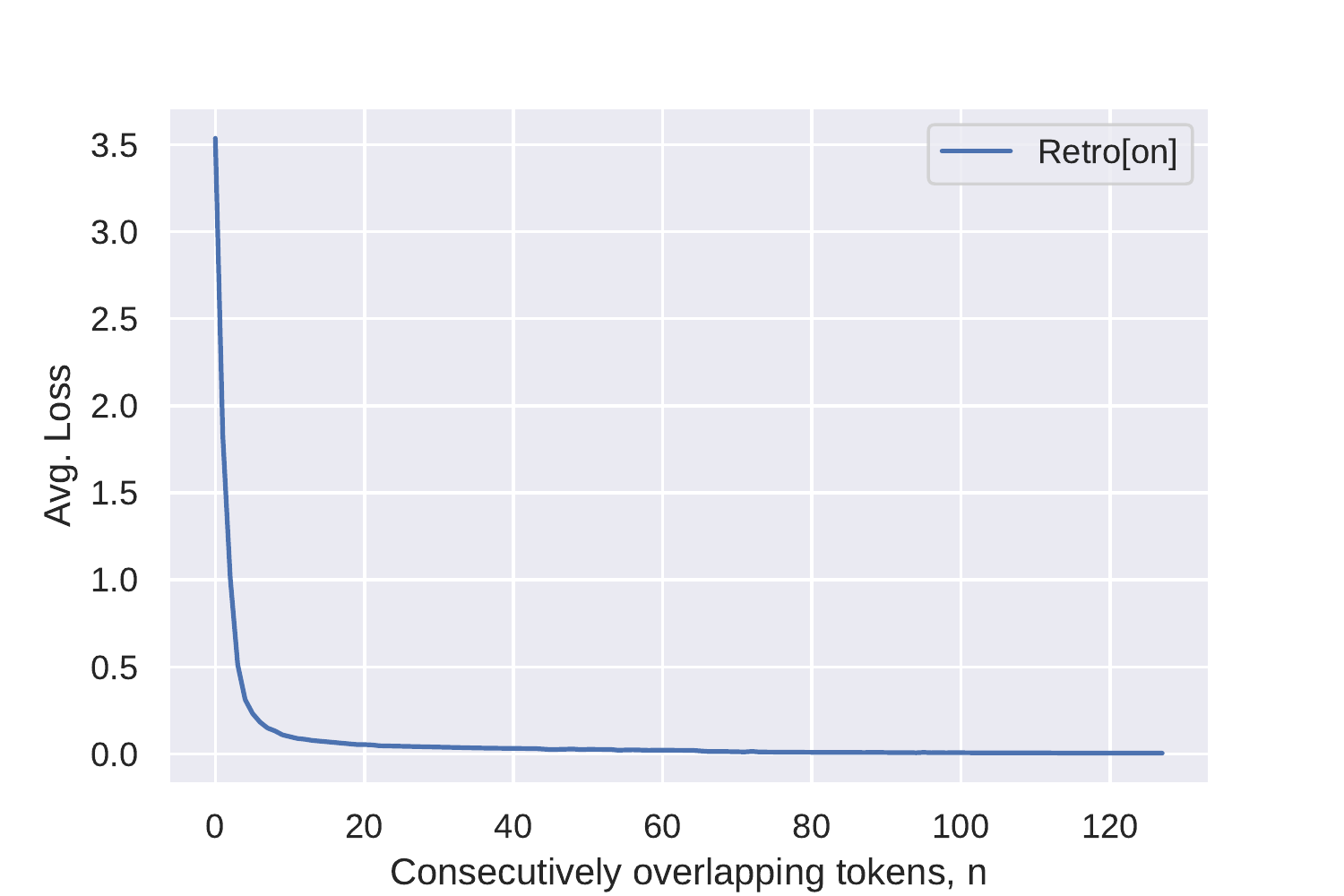}
    \caption{Average loss from \textsc{Retro[on]} over tokens in $\Phi(n)$. Note the drastic decrease with increasing overlap.}
    \label{fig:consecutive_overlap_loss}
\end{figure}

In Figure~\ref{fig:consecutive_overlap_loss} we plot the average loss per bucket,
\begin{equation}
    \frac{1}{|\Phi(n)|}\sum_{x_i\in\Phi(n)} {\mathcal{L}}_{x_i}^{\RETROON}\,,
\end{equation}
as a function of $n$.
Here, ${\mathcal{L}}_{x_i}^{\RETROON}$ is the loss when predicting token $x_i$ using \RETROON\footnote{The sizes of each bucket (accumulated over the validation data) are shown in the appendix, Figure~\ref{fig:overlap_frequencies}.}.
We see that the loss drastically decreases as the consecutive overlap increases.
For example, at an overlap of $n=5$ tokens, the loss is only 6\% of the loss for non-overlapping tokens.
This suggests that \RETRO\ enters ``copy mode'' when the previous tokens overlap with those from a neighbor.

\begin{figure*}[]
    \centering
    \includegraphics[width=\textwidth]{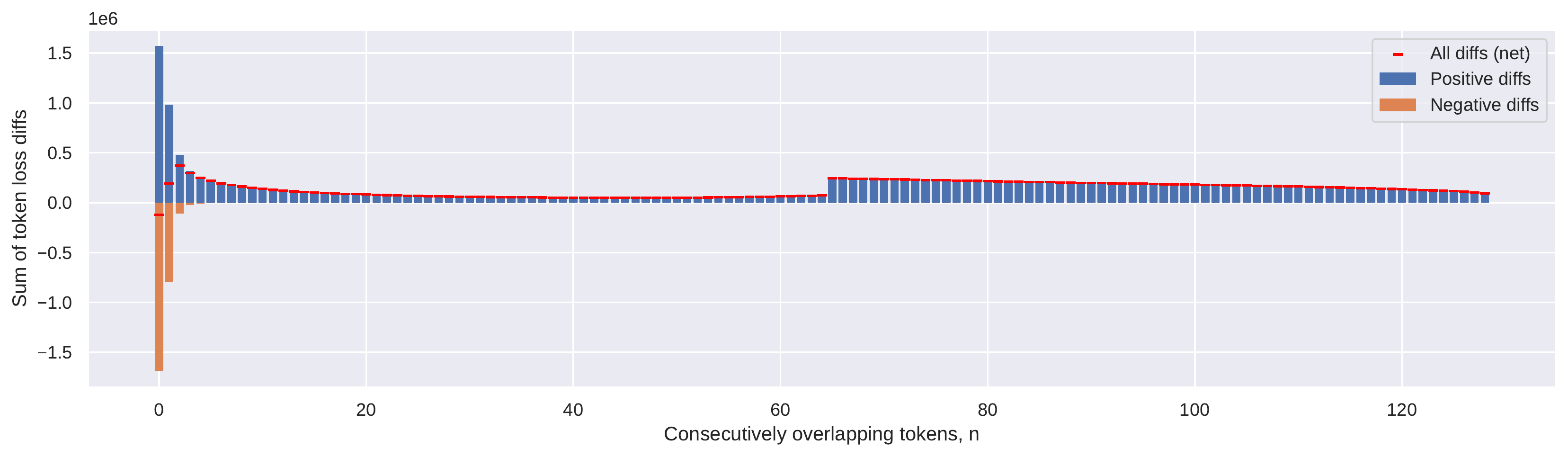}
    \caption{Token-specific loss differences, as distributed over different degrees of overlap. \emph{Positive diffs} shows the sum of all positive loss differences, $\sum_{x_i\in\Phi(n)} \text{max}(0, \Delta \mathcal{L}_{x_i})$, and \emph{Negative diffs} shows the sum of negative loss differences, $\sum_{x_i\in\Phi(n)} \text{min}(0, \Delta \mathcal{L}_{x_i})$. \emph{All diffs} shows the total sum. We see that the vast majority of loss reductions comes from overlapping tokens, e.g.\ $n>0$.}
    \label{fig:total_loss_gains}
\end{figure*}

\subsection{Loss Reductions per Degree of Overlap}
\label{sec:share-of-loss}

For a more detailed analysis of the effect of overlap on predictive performance, we look at the token-specific loss differences between the two configurations \RETROOFF\ and \RETROON:
\begin{align*}
    \Delta \mathcal{L}_{x_i} = \mathcal{L}_{x_i}^{\RETROOFF} - \mathcal{L}_{x_i}^{\RETROON}
\end{align*}
Note that a loss difference $\Delta \mathcal{L}_{x_i}$ is positive if the access to the retrieved context reduces the token-specific loss for~$x_i$.
The overall reduction in loss visible in Figure~\ref{fig:ppl} is the average of the loss differences across all tokens in the validation data.
By aggregating loss differences per bucket $\Phi(n)$, we get a fine-grained picture of how the reductions are distributed with respect to different degrees of consecutive overlap.
This is illustrated in Figure \ref{fig:total_loss_gains}.

For non-overlapping tokens ($n=0$), we can see that there are both positive and negative differences, with a small negative net.
For all overlapping tokens ($n>0$), the net differences are positive, and for buckets with 3 or more overlapping tokens, there are almost no negative differences at all.\footnote{We note a sudden increase in accumulated loss difference for $n>64$ which is expected considering the way in which we construct the buckets; see Appendix \ref{appendix:consecutively_overlapping_tokens} for more details.}
This shows that the largest share of all loss reductions originates from tokens that are consecutively overlapping in neighbors.
Interestingly, the net differences are positive even for very small degrees of overlap.
\citet{retro} considered reductions in bits-per-byte from chunks with up to 8 consecutively overlapping tokens as evidence of a non-trivial generalization capacity.
However, our results suggest that even a small number of overlapping tokens may cause a large reduction in loss, which we take as an argument against this conclusion.

\section{Related Work}

Equipping language models with a retrievable external memory has been extensively studied \citep{guu2020retrieval, karpukhin-etal-2020-dense, lewis2020retrievalaugmented, izacard-grave-2021-leveraging, li2022survey}.
Explicitly leveraging the training data through retrieval to reduce perplexity is proposed in kNN-LM \citep{khandelwal20generalization}. 
kNN-LM matches the leftward context with the leftward context of all training data tokens, and explicitly interpolates between generating and copying the next token.
A recent study analyzes kNN-LM to better understand the causes of performance gains \citep{why_kNNLM_work}.
Similar to our findings in \RETRO{}, lexical overlap has also been found to play a significant role in explaining retrieval performance gains in kNN-LM as well \citep{drozdov2022you}.
The idea of kNN-LM is extended in \textsc{Spalm} \citep{yogatama-etal-2021-adaptive} to instead learn a gating function that facilitates more dynamic interpolation.

In both kNN-LM and \textsc{Spalm}, retrieval is incorporated at the top of the network. 
This might induce a bias towards surface-level rather than semantic augmentation.
In contrast, retrieval in \RETRO\ is incorporated in lower layers of the network, which opens up for more sophisticated integration of the retrieved information. 
Our results suggest, however, that retrieval in \RETRO\ also contributes at the surface rather than at the semantic level, similar to the previous works.

\section{Conclusions and Future Work}

The capacity of language models for generalization is often measured intrinsically using perplexity, loss or bits-per-byte on held-out validation data.
Low perplexity language models perform well as few-shot learners on many downstream tasks due to their capacity to both memorize and non-trivially combine textual information from many sources \citep{gpt3, gopher, J1WhitePaper, palm}.
The hope is that we can externalize memory to reduce the footprints of language models without reducing generalization and downstream task performance.

Our results show that the low loss in \textsc{Retro} almost exclusively originates from tokens overlapping between retrieval and validation data, rather than from more sophisticated generalization.
To better understand this effect, it would be interesting to modify the retrieval component and deliver semantically similar but lexically different context during training.
If the retrieved context is uninformative, the model will learn to ignore it, but if the context is too specific (e.g.\ literal overlap) the model will learn to copy.
By better balancing between these two modes, models may become better at utilizing retrieved information at a deeper and more generalizable level.

\section*{Limitations}

We have made our best effort in trying to reproduce the model and results of \citet{retro}.
Nonetheless, our experiments were performed on one of the smaller model sizes and with a dataset that is only ${\sim}$2.5\% of their size (52 billion vs.\ 2 trillion tokens).
This was due to computational constraints and lack of larger open datasets.
However, as was also shown by \citet{retro}, the performance gain of retrieval is constant with respect to model size.
We speculate that larger \RETRO\ models mostly improve with respect to loss on tokens that are not overlapping, which would not change our conclusions here.

One noteworthy limitation of our work is the fact that we compare to a non-retrieval baseline (\textsc{Retro[off]}) that was trained with access to retrieved context.
We were not able to train a separate non-retrieval baseline due to computational constraints, but note that the bits-per-byte results of \textsc{Retro[off]} and the baseline in \citet{retro} were close to identical.

\section*{Acknowledgements}
This work was partially supported by the Wallenberg AI, Autonomous Systems and Software Program (WASP) funded by the Knut and Alice Wallenberg Foundation.
The computations were enabled by resources provided by the National Academic Infrastructure for Supercomputing in Sweden (NAISS) at Alvis partially funded by the Swedish Research Council through grant agreement no. 2022-06725, and by the Berzelius resources provided by the Knut and Alice Wallenberg Foundation at the National Supercomputer Centre.

\bibliography{anthology,custom}
\bibliographystyle{acl_natbib}

\appendix

\newpage
\clearpage

\begin{minipage}{0.98\textwidth}
\begin{center}
  \includegraphics[width=0.95\textwidth]{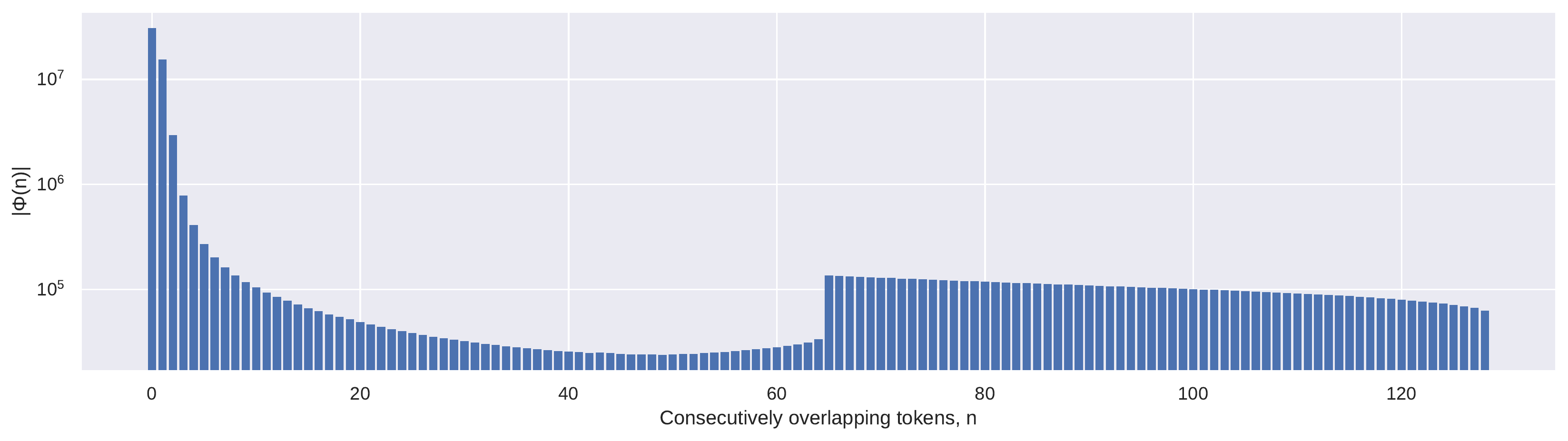}
\captionof{figure}{
Number of validation set tokens in each bucket $\Phi(n)$. Since the neighbors have a maximal length of 128 tokens, this is also the longest possible overlap $n$.
}\label{fig:overlap_frequencies}
\end{center}

\vspace{1mm}

\begin{center}
\begin{tabular}{lllll}
\hline
                            &              & \textbf{Documents} &   \textbf{Chunks} & \textbf{Tokens} \\ \hhline{=====}
\multirow{5}{*}{Training}   & Pile-CC      &   15,728k &     269M &  16.7B \\
                            & Wikipedia En &    5,082k &      61M &   3.8B \\
                            & GitHub       &    5,417k &     181M &  11.4B \\
                            & Books3       &       83k &     191M &  12.2B \\
                            & RealNews     &    9,360k &     130M &   8.0B \\ 
                            & \textbf{Total}        &   35,670k &     833M &  52.2B \\ \hline
\multirow{5}{*}{Validation} & Pile-CC      &     52.8k &   900.4k &  56.0M \\
                            & Wikipedia En &     17.4k &   215.9k &  13.3M \\
                            & GitHub       &     18.3k &   598.4k &  37.7M \\
                            & Books3       &      0.3k &   727.6k &  46.5M \\
                            & RealNews     &     16.4k &   234.5k &  14.5M \\
                            & \textbf{Total}        &    105.3k & 2,676.8k & 168.0M \\ \hline
\end{tabular}
\captionof{table}{
Statistics for our MassiveOpenText dataset. We use the web text, Wikipedia, GitHub and Books3 corpora from the Pile, and news text from RealNews.
}\label{tab:massiveopentext}

\end{center}

\end{minipage}


\vspace{1mm}
\section{MassiveOpenText statistics}
\label{sec:appendix}

Statistics on the number of documents, chunks and tokens for each split and text category are shown in Table \ref{tab:massiveopentext}.

\section{\RETRO{}  model details}
We show hyperparameters of our \RETRO{} model in Table \ref{tab:models}.

\begin{table}[htbp]
\begin{tabular}{lllll}
\hline
                         & \textbf{Param} &         \\ \hhline{===}
\multirow{5}{*}{Encoder} & Num layers  &          2 \\
                         & Num heads   &         14 \\
                         & Hidden size &        896 \\
                         & FFN         &       3584 \\
                         & CA layers   &        [2] \\ \hline
\multirow{5}{*}{Decoder} & Num layers  &         12 \\
                         & Num heads   &         12 \\
                         & Hidden size &       1536 \\
                         & FFN         &       6144 \\
                         & CCA layers  &   [6,9,12] \\ \hline
\end{tabular}
\caption{\label{tab:models} Hyperparameters of our trained Retro model.}
\end{table}

\newpage
\mbox{~}
\vspace{130mm}
\section{Consecutively overlapping tokens}
\label{appendix:consecutively_overlapping_tokens}

As explained in Section \ref{sec:consecutive_overlap}, we sort validation set tokens into buckets denoted $\Phi(n)$ depending on the longest overlapping leftward context.

In Figure \ref{fig:overlap_frequencies} we show the number of tokens in each bucket.
We note a big ``jump'' from $n=64$ to $n=65$, which can be explained by the following rationale.
A neighbor $[N, F]$ to a chunk $C_i$ is retrieved based on the similarity between $C_i$ and $N$.
In the case where both $C_i = N$ and $C_{i+1} = F$, tokens in $C_{i+1}$ will be put into $\Phi(n)$ with $n=65, \dots, 128$.
The jump in Figure \ref{fig:overlap_frequencies} indicates such duplicates are common in our data.

\section{Model validation}
\label{appendix:model_validation}
As we aim to reproduce the 425M model trained in \citet{retro}, it is important to validate that the implementations are equivalent and that their evaluation results are comparable. 
However, evaluations of the 425M model in \citet{retro} on the Pile are not available, making it hard to make direct comparisons.
\citet{retro} report evaluation results on the C4 \citep{T5} dataset, with various sizes of retrieval datasets.
For their setup with 36B retrieval tokens, which is the most similar to our own retrieval size, they report that bits-per-byte is reduced by $\sim$ 2\% (from 0.92 to 0.90) when using retrieval. 
That could be compared to our results on Pile-CC, as both datasets originate from Common Crawl.
In our experiments, loss is reduced by 7\% (from 3.05 to 2.83) on Pile-CC.

Evaluations on the Pile in \citet{retro} are only reported for their largest model (7B params) and largest retrieval set (2T tokens). 
For example, on Pile–GitHub their reduction is ${\sim}$53\% whereas our reduction is 42\%.

While these numbers are not directly comparable, we believe they indicate that our reimplementation of the \RETRO\ model is working as expected.

\end{document}